\documentclass[letterpaper]{article} % DO NOT CHANGE THIS
\usepackage{aaai2026}  % DO NOT CHANGE THIS
\usepackage{times}  % DO NOT CHANGE THIS
\usepackage{helvet}  % DO NOT CHANGE THIS
\usepackage{courier}  % DO NOT CHANGE THIS
\usepackage[hyphens]{url}  % DO NOT CHANGE THIS
\usepackage{graphicx} % DO NOT CHANGE THIS
\usepackage{natbib}  % DO NOT CHANGE THIS AND DO NOT ADD ANY OPTIONS TO IT
\usepackage{caption} % DO NOT CHANGE THIS AND DO NOT ADD ANY OPTIONS TO IT
\usepackage{algorithm}
\usepackage{algorithmic}
\usepackage{booktabs}
\usepackage{multirow}
\usepackage{pifont} 
\usepackage{amsmath}
\usepackage{subcaption}
\usepackage{amssymb}
\graphicspath{{Img/}}
\usepackage{newfloat}
\usepackage{listings}
\DeclareCaptionStyle{ruled}{labelfont=normalfont,labelsep=colon,strut=off} % DO NOT CHANGE THIS
\floatstyle{ruled}
\newfloat{listing}{tb}{lst}{}
\floatname{listing}{Listing}
\title{SEMC: Structure-Enhanced Mixture-of-Experts Contrastive Learning for\\ Ultrasound Standard Plane Recognition}
\author{
    Qing Cai\textsuperscript{\rm 1},\
    GuihaoYan\textsuperscript{\rm 1},\
    Fan Zhang\textsuperscript{\rm 2}\thanks{Corresponding authors.},\
    Cheng Zhang\textsuperscript{\rm 1}$^*$,\
    Zhi Liu\textsuperscript{\rm 3}
}
\affiliations{
    \textsuperscript{\rm 1}Faculty of Information Science and Engineering, Ocean University of China\\
    \textsuperscript{\rm 2}Department of Health Technology and Informatics, The Hong Kong Polytechnic University\\
    \textsuperscript{\rm 3}School of Information Science and Engineering, Shandong University\\
    cq@ouc.edu.cn, \{yanguihao, zhangcheng\}@stu.ouc.edu.cn, fan-hti.zhang@polyu.edu.hk, liuzhi@sdu.edu.cn
}

\usepackage{bibentry}
\begin{document}

\maketitle

\begin{abstract}
Ultrasound standard plane recognition is essential for clinical tasks such as disease screening, organ evaluation, and biometric measurement. However, existing methods fail to effectively exploit shallow structural information and struggle to capture fine-grained semantic differences through contrastive samples generated by image augmentations, ultimately resulting in suboptimal recognition of both structural and discriminative details in ultrasound standard planes. To address these issues, we propose SEMC, a novel \underline{S}tructure-\underline{E}nhanced \underline{M}ixture-of-Experts \underline{C}ontrastive learning framework that combines structure-aware feature fusion with expert-guided contrastive learning. Specifically, we first introduce a novel Semantic-Structure Fusion Module (SSFM) to exploit multi-scale structural information and enhance the model's ability to perceive fine-grained structural details by effectively aligning shallow and deep features. Then, a novel Mixture-of-Experts Contrastive Recognition Module (MCRM) is designed to perform hierarchical contrastive learning and classification across multi-level features using a mixture-of-experts (MoE) mechanism, further improving class separability and recognition performance. More importantly, we also curate a large-scale and meticulously annotated liver ultrasound dataset containing six standard planes. Extensive experimental results on our in-house dataset and two public datasets demonstrate that SEMC outperforms recent state-of-the-art methods across various metrics.
\end{abstract}

\begin{links}
    \link{Code}{https://github.com/YanGuihao/SEMC}
    \link{Datasets}{https://github.com/YanGuihao/SEMC}
\end{links}

\section{Introduction}
\begin{figure}[!t]  
  \centering
  \includegraphics[width=\columnwidth,height=1.3\columnwidth]{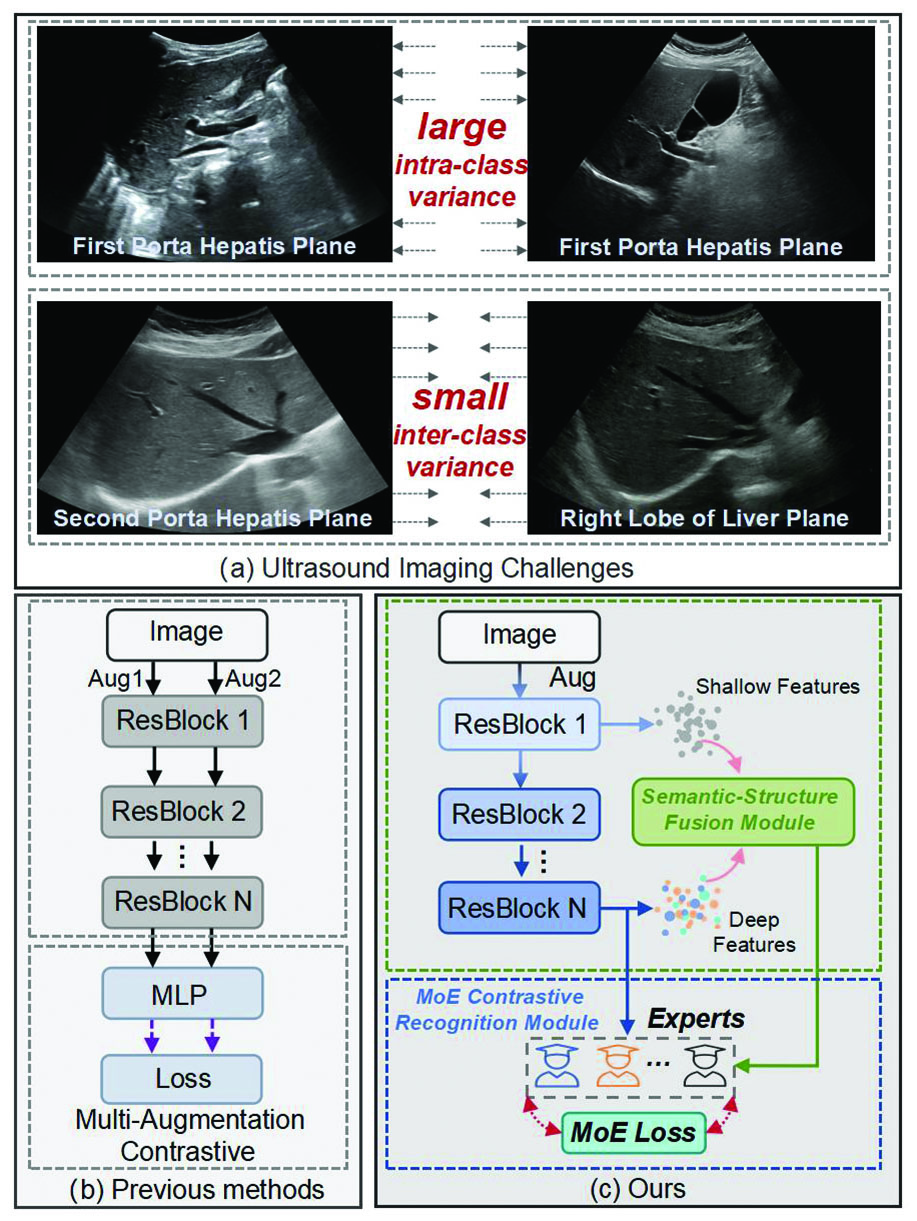} 
\caption{
(a) Ultrasound standard planes exhibit large intra-class variance, where images from the same plane can appear markedly different, and small inter-class variance, where different planes often share highly similar visual patterns. (b) Previous methods mainly rely on deep semantic features, neglecting shallow structural cues. (c) In contrast, Our SEMC framework integrates the shallow structure via semantic-structure fusion and employs a MoE for hierarchical contrastive learning,  which ca yields more discriminative and structure-aware representations.
}
  \label{fig1}
\end{figure}
Ultrasound imaging is one of the most widely used medical imaging techniques in clinical practice, owing to its non-invasive nature, real-time capability, high efficiency, and low cost~\cite{spencer2008utility}. It is particularly effective for visualizing human organs and soft tissues and is widely used in prenatal examinations. In clinical workflows, acquiring standard planes (SP) is essential for accurate diagnosis and quantitative assessment. These planes provide clinicians with reliable structural visualization and consistent anatomical reference points for measurement~\cite{wang2022task,di2022deep}. For instance, in prenatal ultrasound, the femoral standard plane, thalami standard plane, and abdominal standard plane are commonly used to measure fetal length, head circumference, and abdominal circumference, respectively. These biometric measurements serve as key indicators for evaluating fetal growth~\cite{salomon2006feasibility,guo2022fetal}. However, the quality of acquired standard planes can vary considerably depending on the operator’s experience and scanning technique, which may influence the accuracy of growth assessments and subsequent clinical decisions~\cite{salomon2011practice,maraci2014searching}.

Recent studies have explored deep learning-based methods for standard plane (SP) identification and have achieved promising results~\cite{pu2021automatic,migliorelli2024use}. However, SP recognition still face several critical challenges. The quality of ultrasound images varies considerably due to speckle noise, low contrast, and indistinct anatomical boundaries, making structural region detection inherently difficult. As shown in Figure~\ref{fig1}(a), images from the same anatomical plane show substantial appearance variations caused by inconsistent acquisition angles, probe pressure, and operator experience~\cite{lin2019multi,xie2020using,yu2024lpc,krishna2024standard}, while images from distinct planes often exhibit subtle visual differences due to low contrast and ambiguous boundaries, requiring fine-grained discrimination~\cite{baumgartner2016real,baumgartner2017sononet}. Most existing approaches focus primarily on deep semantic representations while overlooking shallow structural cues~\cite{cai2018adaptive,zhang2024cross,yan2025sgtc}, thereby limiting the model’s ability in both semantic discrimination and structural perception~\cite{men2023gaze,li2025rule,Liu2024,zhang2024federated}. Moreover, although contrastive learning has been incorporated as an auxiliary strategy by constructing augmented positive and negative pairs, these techniques often struggle to capture the fine-grained semantic distinctions inherent to ultrasound images, as illustrated in Figure~\ref{fig1}(b).

In response to these gaps,  we propose a novel structure-enhanced mixture-of-experts contrastive learning framework, dubbed SEMC, which effectively integrates structure-aware feature fusion with expert-guided contrastive learning to tackle the challenges inherent in ultrasound standard plane recognition. Specifically, in this framework, we design a novel Semantic-Structure Fusion Module (SSFM) that explicitly aligns and integrates shallow structural cues with deep semantic representations. It enhances the model's sensitivity to fine-grained structural details. To further enhance the model’s discriminative capability, we designs a new Mixture-of-Experts Contrastive Recognition Module (MCRM), in which multiple expert branches are specifically designed to specialize in different aspects of the feature space and collaboratively perform hierarchical contrastive learning, as illustrated in Figure~\ref{fig1}(c). By enforcing contrastive objectives at multiple feature levels, the framework promotes improved inter-class separability and more compact intra-class clustering within the representation space. Additionally, we construct a high-quality liver ultrasound dataset, \textbf{LP2025}, containing six standard planes to address the scarcity of publicly available data for standard plane recognition. Evaluations on this dataset and two public standard plane benchmark datasets demonstrate that SEMC outperforms existing state-of-the-art methods across multiple metrics, showing strong potential for clinical application. In summary, our main contributions are as follows:

\begin{itemize}
    \item We introduce a novel structure-enhanced mixture-of-experts contrastive learning framework, dubbed SEMC, which integrates the semantic-structure fusion and MoE contrastive recognition modules to enhance fine-grained structural perception and discriminative feature representation for ultrasound plane recognition.  
    \item We construct LP2025, a high-quality liver ultrasound dataset comprising six standard planes, addressing the scarcity of publicly available benchmarks and supporting further research in standard plane recognition.
    \item Extensive experiments on two public datasets and our in-house liver ultrasound dataset demonstrate that SEMC framework consistently outperforms existing state-of-the-art methods in standard plane recognition tasks.

\end{itemize}

\section{Related Work}

\subsection{Standard Plane Recognition in Ultrasound}

Standard plane recognition is a fundamental task in medical image understanding, with broad clinical applications such as disease screening, organ function assessment, and biometric measurement. Early methods relied on handcrafted features combined with traditional classifiers (\textit{e.g.}, SVM and KNN), but their generalization capability was limited due to weak feature representation and low quality of ultrasound~\cite{christodoulou2003texture,latha2020carotid,huang2020texture,liao2024machine}. In recent years, convolutional neural networks (CNNs) have become the mainstream solution. For example, the SonoNet~\cite{baumgartner2017sononet} series, built on the VGG architecture, achieved promising performance in fetal standard plane recognition. Subsequent studies have incorporated multi-task learning, attention mechanisms~\cite {cai2021avlsm,zhang2024exploring}, and structural priors to enhance the model’s ability to identify key regions and capture fine-grained variations~\cite {cai2018multi,zhu2022gaze,yu2024lpc,ciobanu2025automatic}. However, these methods mainly rely on high-level features and often overlook shallow structural cues and spatial context, leading to degraded performance under subtle inter-class differences or complex imaging conditions. To overcome this, we introduce a semantic–structure fusion module that integrates shallow and deep features to enhance structural representation and discrimination.

\begin{figure*}[!t]
	\centering
	\includegraphics[width=\textwidth,height=0.4\textheight]{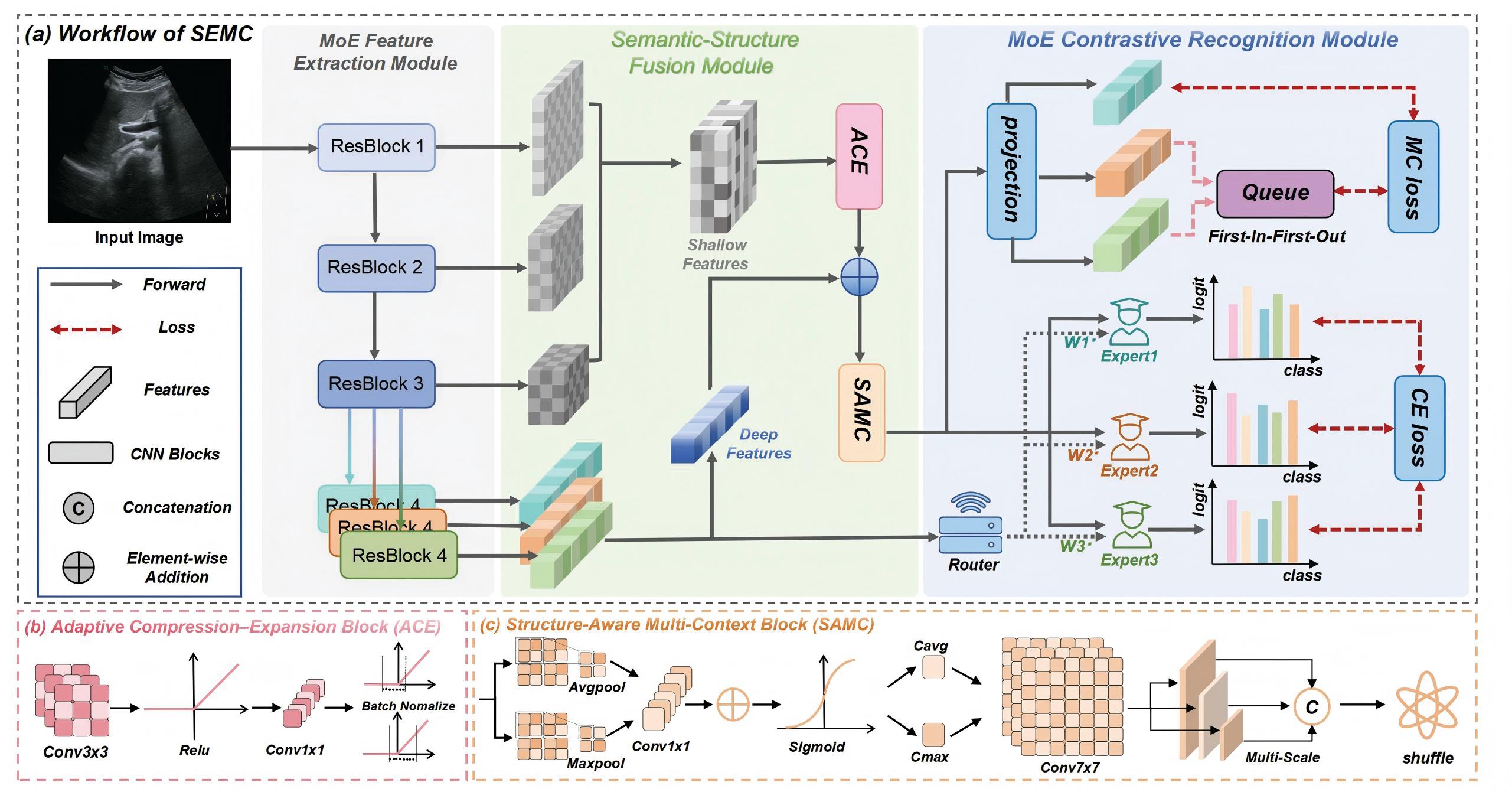} 
	\caption{Architecture of the proposed SEMC framework. The framework first employs an MoE-based feature extractor to generate multi-level expert features from the input ultrasound image. These features are then aligned and enhanced through a Semantic-Structure Fusion Module (SSFM). The resulting representations are fed into the Mixture-of-Experts Contrastive Recognition Module (MCRM), which consists of two branches: a multi-class classification headserving as the primary task, and an MoE-based contrastive learning branch serving as an auxiliary task to further improve the primary task by refining the learned feature representations.}
	\label{fig2}
\end{figure*}
\subsection{Contrastive Learning and Mixture-of-Experts}
Contrastive learning has shown strong potential for improving representation learning, particularly in medical imaging tasks where data are limited and class boundaries are ambiguous. Methods such as MoCo~\cite{he2020momentum} and SimCLR~\cite{chen2020simpleframeworkcontrastivelearning} optimize the representation space by constructing positive and negative sample pairs, promoting intra-class compactness and inter-class separability. Supervised contrastive learning~\cite{khosla2021supervisedcontrastivelearning,lin2024consistent} further enhances discriminability and semantic consistency by leveraging label information during pair construction. Meanwhile, the MoE paradigm has gained increasing attention for its dynamic modeling capabilities and parameter efficiency~\cite{shazeer2017outrageouslylargeneuralnetworks,riquelme2021scaling,zoph2022designing}. For instance, Conditional MoE~\cite{zhu2022uniperceivermoe} and Switch Transformer~\cite{fedus2022switchtransformersscalingtrillion} have achieved substantial breakthroughs in both natural language processing and computer vision. Nevertheless, in ultrasound image analysis, where anatomical structures are complex and boundaries often indistinct, the integration of MoE with contrastive learning remains underexplored. Existing methods lack effective mechanisms to guide expert collaboration using structural cues. To this end, we propose a framework that combines structure-enhanced feature fusion with contrastive expert modeling,explicitly improving the model’s ability to recognize fine-grained differences in standard plane classification tasks.

\section{Methodology}
\subsection{Workflow Overview}
We propose a novel Structure-Enhanced Mixture-of-Experts Contrastive (SEMC) learning framework for ultrasound standard plane recognition. As illustrated in Figure~\ref{fig2}, our core contributions include two main components: (1) the semantic-structure fusion module, which explicitly aligns and integrates multi-level features to enhance structural awareness, and (2) the MoE-based contrastive recognition module, which leverages expert-specific features for both contrastive learning and classification. By jointly optimizing the feature space of supervised and self-supervised learning, this module significantly improves the class separability and recognition performance of the model. In the following sections, we will provide the details of these two modules as well as our in-house dataset.
\section{}

\subsection{LP2025 Dataset Construction}
\noindent \textbf{Data Collection}. To advance research on standard plane recognition in ultrasound imaging, we introduce LP2025, a comprehensive and high-quality dataset specifically curated for deep learning-based anatomical understanding and classification. The dataset was developed under the clinical supervision of experienced radiologists and certified sonographers from a leading tertiary medical center. All ultrasound scans were acquired using high-resolution diagnostic systems to ensure excellent image quality and clear structural visibility. A standardized imaging protocol was strictly followed to harmonize scanning procedures across different patients and sessions, ensuring consistency in anatomical coverage, spatial resolution, and diagnostic relevance.

Each subject underwent a systematic abdominal ultrasound examination, during which six clinically meaningful liver standard planes were meticulously captured. These planes were selected based on their diagnostic relevance in hepatobiliary evaluations and their frequent usage in routine clinical workflows. The six standard planes in LP2025 are as follows:

\begin{itemize}
    \item First Porta Hepatis Plane (FHP1): Captures the bifurcation of the portal vein, serving as a key landmark for hepatic segmentation.
    \item Second Porta Hepatis Plane (FHP2): Displays the continuation of the portal vein and hepatic artery, facilitating vascular assessments.
    \item Left Lobe Plane (LLP): Highlights the morphology and parenchymal pattern of the left hepatic lobe.
    \item Right Lobe Plane (RLP): Visualizes the texture and size of the right hepatic lobe, often used to assess hepatomegaly and hepatic lesions.
    \item Sagittal Plane of Left Portal Vein (LPV-S): Offers a clear sagittal view of the left portal vein branch, aiding in vascular diagnosis.
    \item Hepatorenal Plane (HRP): Shows the interface between the liver and right kidney, commonly used to detect ascites or space-occupying lesions.
\end{itemize}

In addition to the six standard planes, LP2025 includes a Non-Standard Plane (NSP) category, comprising images that do not correspond to the above-defined diagnostic views but are frequently encountered in routine ultrasound examinations. The inclusion of this category introduces realistic variability and classification ambiguity, thereby enhancing model robustness and better reflecting real-world clinical deployment scenarios.

\begin{table}[!t]
	\centering
	\footnotesize
	\setlength{\tabcolsep}{3.8pt}
	\setlength{\tabcolsep}{1.0mm}{
		\begin{tabular}{lcc}
			\hline\specialrule{0.06em}{0pt}{1.9pt}
			\textbf{Plane} & \textbf{Abbreviation} & \textbf{Number} \\
			\specialrule{0.02em}{0pt}{1.9pt}
			First Porta Hepatis Plane & FHP1 & 979 \\
			Second Porta Hepatis Plane & FHP2 & 324 \\
			Left Lobe Plane & LLP & 1038 \\
			Right Lobe Plane & RLP & 490 \\
			Sagittal Plane of Left Portal Vein & LPV-S & 840 \\
			Hepatorenal Plane & HRP & 1072 \\
			Non-Standard Plane & NSP & 4626 \\
			\specialrule{0.06em}{0pt}{1.9pt}
			\textbf{Total} & -- & 9369 \\
			\hline\specialrule{0.06em}{0pt}{1.9pt}
	\end{tabular}}
	\caption{Image distribution of the LP2025 dataset across six liver standard planes and the non-standard (NSP) category.}
	\label{tab:lp2025_stats}
\end{table}
\subsection{Dataset Composition and Annotation Quality}

To ensure the accuracy, consistency, and clinical validity of the labels, each image in the LP2025 dataset was independently annotated by a team of senior sonographers, all of whom have more than five years of hands-on experience in liver ultrasound imaging. The annotation process focused on two key aspects: standard plane classification and the presence of clearly identifiable anatomical structures.

A rigorous multi-stage quality control pipeline was implemented to maintain high annotation standards:

\begin{itemize}
    \item Initial Review: Each annotation was independently cross-checked by two sonographers to detect potential inconsistencies or errors.
    \item Consensus Verification: For cases with disagreement, at least three senior sonographers engaged in a consensus discussion to ensure clinically reliable labels.
    \item Final Validation: A final round of inspection was performed to assess the clinical relevance of each image and to exclude low-quality or ambiguous samples that could negatively influence model training or evaluation.
\end{itemize}

This comprehensive, multi-expert review process ensures the reliability and trustworthiness of the LP2025, establishing a robust foundation for both algorithm development and clinically oriented research.

Table~\ref{tab:lp2025_stats} summarizes the LP2025 dataset, which contains 9,369 high-quality, clinically validated liver ultrasound images across six standard planes. All patient data were thoroughly anonymized, with no identifiable information retained during collection, processing, or release.

\subsection{Mixture-of-Experts Feature Extraction Module}
To capture fine-grained variations and complex anatomical structures in ultrasound images, we design a MoE feature extraction module. The first three blocks (\textit{i.e.}, \texttt{layer1} to \texttt{layer3}) are shared across all branches and serve as a common encoder for extracting low- and mid-level features. Beyond \texttt{layer3}, we introduce three parallel, structurally identical yet parameter-independent fourth-stage blocks (denoted as \texttt{layer4-1}, \texttt{layer4-2}, and \texttt{layer4-3}), forming three specialized deep expert pathways:
\begin{equation}
F_1, F_2, F_3 = \text{ResNet}_{1\sim3}(x),
\end{equation}
\begin{equation}
D_1 = \text{ResNet}_{4-1}(F_3),
\end{equation}
\begin{equation}
D_2 = \text{ResNet}_{4-2}(F_3),
\end{equation}
\begin{equation}
D_3 = \text{ResNet}_{4-3}(F_3),
\end{equation}
where $F_3$ denotes the output feature of the shared backbone, and $\{D_1, D_2, D_3\}$ represent the high-level semantic features extracted by each expert path. This design introduces diverse feature representations through decoupled expert parameters, enabling the modeling and selection of different semantic perspectives in subsequent fusion modules. 
\subsection{Semantic-Structure Fusion Module (SSFM)}
Most existing methods primarily focus on deep features while overlooking the complementary value of shallow features, particularly in cases where anatomical contrast is weak or boundaries are indistinct. To address this limitation, we propose a novel Structure-Semantic Fusion Module (SSFM), which integrates shallow and deep features through two components, \textit{i.e.}, the Adaptive Compression-Expansion (ACE) Block and the Structure-Aware Multi-Context (SAMC) Block. This design enhances both feature discrimination and structural representation.

\noindent \textbf{Adaptive Compression–Expansion Block (ACE).} 
To address the spatial and channel mismatches between shallow and deep features, we propose a lightweight ACE module. It aligns the shallow features $\{F_1, F_2, F_3\}$ with the deep expert features through progressive downsampling and channel adaptation. Each ACE block processes an input $\mathbf{X}_0 = F_i$ through $L$ sequential stages:
\begin{equation}
\begin{aligned}
 \mathbf{X}_{i+1} &= \mathrm{BN} \Big( \mathrm{Conv}_{1 \times 1} \big( \mathrm{ReLU} \big( \mathrm{BN} \cdot \\ & \quad\quad\quad
 \mathrm{DWConv}_{3 \times 3}^{s=2} (\mathbf{X}_i) \big) \big) \Big), \\&\quad\quad\quad i=0,1,\ldots,L-1,
\end{aligned}
\end{equation}
where the channel size doubles at each step, following $C_i = C_{\mathrm{in}} \times 2^i$. After $L$ stages, a $1\times1$ convolution followed by BN and ReLU maps $\mathbf{X}_L$ to the target channel dimension $C_{\mathrm{out}}$:
\begin{equation}
    F_i' = \mathrm{ReLU} \left( \mathrm{BN} \left( \mathrm{Conv}_{1\times1} (\mathbf{X}_L) \right) \right),
\end{equation}
yielding the aligned features $F_i' \in \mathbb{R}^{C_{\mathrm{out}} \times H_L \times W_L}$.

ACE first reduces spatial resolution using strided depth-wise convolutions, and then adjusts channels dimensions through pointwise convolutions, computational efficiency while preserving structural information. The aligned shallow features are subsequently fused with deep expert features $\{D_1, D_2, D_3\}$ through element-wise addition:
\begin{equation}
    M_i = F_i' + D_i, \quad i = 1,2,3.
\end{equation}

Compared with feature concatenation, this fusion strategy avoids channel redundancy, reduces parameters, and encourages the learning of shared discriminative patterns.

\noindent \textbf{Structure-Aware Multi-Context Block (SAMC).}
To enhance the discriminative power and structural representation of the fused features, we propose a novel SAMC module, which reconstructs feature patterns across multiple receptive fields using a set of parallel multi-scale convolutions. Additionally, a coordinated channel–spatial attention mechanism adaptively highlights informative responses. By jointly modeling semantic cues and spatial structures, the SAMC enables model to capture fine-grained anatomical details while suppressing irrelevant background variations. For each fusion branch \(\mathcal{M}_i\), the processing is performed as follows:
\begin{align}
\mathbf{C}_i = \sigma \Big( 
& \mathrm{FC}_2 \big(\delta(\mathrm{FC}_1(\mathrm{AvgPool}(\mathcal{M}_i))) \big) \nonumber \\
& + \mathrm{FC}_2 \big(\delta(\mathrm{FC}_1(\mathrm{MaxPool}(\mathcal{M}_i))) \big) 
\Big), \\
& \quad\quad\quad\quad\quad\quad\quad\quad\quad i = 1, 2, 3,  \nonumber
\end{align}
where \(\mathbf{C}_i \in \mathbb{R}^C\) is the adaptive channel attention. Global pooling captures context, while shared  fully connected layers with activation \(\delta(\cdot)\) capture inter-channel dependencies. The sigmoid \(\sigma(\cdot)\) produces normalized attention weights.
\begin{align}
    \mathbf{S}_i &= \sigma \left( \mathrm{Conv} \left( 
    \left[ \mathrm{Mean}(\mathbf{C}_i \odot \mathcal{M}_i, \mathrm{dim}=1), \right. \right. \right. \nonumber \\
    & \quad \left. \left. \left. \mathrm{Max}(\mathbf{C}_i \odot \mathcal{M}_i, \mathrm{dim}=1) \right] \right) \right), \quad i = 1, 2, 3,
\end{align}
where $\mathbf{S}_i \in \mathbb{R}^{1 \times H \times W}$ denotes the spatial attention map generated from the channel-refined features. A convolutional layer aggregates spatial cues and guides the network to emphasize anatomically relevant regions.
\begin{align}
\mathbf{O}_i &= \mathrm{Conv} \Big( \mathrm{Shuffle} \big( \mathrm{Concat}(\{\mathbf{F}_k^{(i)}\}_{k=1}^K) \big) \Big), \\
& \quad\quad\quad\quad\quad\quad\quad\quad\quad\quad\quad\quad i = 1, 2, 3, \nonumber
\end{align}
where \(\{\mathbf{F}_k^{(i)}\}_{k=1}^K\) denote the multi-scale features extracted from spatially enhanced input \(\mathbf{S}_i \odot \mathbf{C}_i \odot \mathcal{M}_i\). These features are concatenated, channel-shuffled to facilitate cross-channel interaction, and compressed via pointwise convolution to produce the fused output \(\mathbf{O}_i\).

\begin{table*}[t]
\footnotesize
\centering
\setlength{\belowcaptionskip}{0cm}
\renewcommand{\arraystretch}{1.0}
\begin{centering}
	\setlength{\tabcolsep}{1.8mm}{
\begin{tabular}{lcccccccc}
\hline\specialrule{0.06em}{0pt}{1.9pt}
\multicolumn{1}{c}{}&\multicolumn{4}{c}{\textbf{FPUS23}} & \multicolumn{4}{c}{\textbf{CAMUS}} \\\cmidrule(lr){2-9}
Method & Accuracy$\uparrow$ & Precision$\uparrow$ & Recall$\uparrow$ & F1-score$\uparrow$ 
& Accuracy$\uparrow$ & Precision$\uparrow$ & Recall$\uparrow$ & F1-score$\uparrow$ \\ 

\specialrule{0.02em}{0pt}{1.9pt}
Diffmic~\cite{yang2023diffmicdualguidancediffusionnetwork}        & 95.29 & 80.63 & 81.58 & 81.08 & 80.91 & 80.25 & 79.20 & 79.69 \\
Area~\cite{chen2023area}           & 95.20 & 93.37 & 95.71 & 94.40 & 81.59 & 80.17 & \textbf{82.53} & 80.88 \\
Shike~\cite{jin2023long}          & 95.15 & 93.80 & 94.93 & 94.31 & 80.48 & 79.58 & 79.79 & 79.52 \\
Metaformer~\cite{yu2023metaformer}     & 95.52 & 94.19 & 94.48 & 94.53 & 81.52 & 81.58 & 79.69 & 80.49 \\
Cast~\cite{ke2024learninghierarchicalimagesegmentation}           & 95.24 & 94.23 & 93.99 & 94.43 & 81.34 & 80.85 & 79.88 & 80.31 \\
Supmin~\cite{mildenberger2025taleclassesadaptingsupervised}         & 95.28 & 94.03 & 94.68 & 94.34 & 81.13 & 80.88 & 78.84 & 79.71 \\
SEMC (Ours)           & \textbf{95.78} & \textbf{94.38} & \textbf{95.81} & \textbf{95.06} & \textbf{82.13} & \textbf{82.03} & 80.08 & \textbf{80.93} \\
\hline\specialrule{0.06em}{0pt}{1.9pt}
\end{tabular}}

\caption{Quantitative comparisons of different models on the FPUS23 and CAMUS datasets.}
\label{tab:FPUS2023andCamus}
\end{centering}
\end{table*}

\subsection{MoE Contrastive Recognition Module (MCRM)}
Existing methods have introduced contrastive learning as an auxiliary recognition strategy by constructing augmented positive and negative sample pairs during training. However, these approaches often struggle to effectively capture the inherent fine-grained semantic variations in ultrasound images. To address this, we designs a novel MoE contrastive recognition module, which consists of two synergistic branches. (1) MoE Contrastive Branch: Multiple expert subnetworks focus on different regions of the feature space and collaboratively perform hierarchical contrastive learning. This enhances the inter-class separability and intra-class compactness in the learned representations. (2) MoE Recognition Branch: Expert subnetworks extract complementary discriminative cues from diverse semantic perspectives and spatial scales, thereby improving the model’s ability to accurately recognize various standard planes.

\noindent \textbf{MoE Contrastive Branch.}
To fully exploit multi-level semantic and spatial information for ultrasound plane recognition, we propose a MoE-Enhanced Contrastive Branch comprising three expert branches that share a common backbone but are supervised by different fusion views. The first expert output \(\mathbf{O}_1\) is used as the contrastive anchor (query), whereas \(\mathbf{O}_2\) and \(\mathbf{O}_3\) serve as positive keys for updating a dynamic queue that supports negative sampling. Classification logits from all experts are concatenated, and the ground-truth labels are replicated for semantic supervision. The current expert features are further integrated with a momentum memory queue \(\mathcal{Q}\), which stores historical representations to facilitate structural contrastive learning:
\begin{align}
\mathbf{O}_{\text{con}} &= \mathrm{Concat}(\mathbf{O}_1, \mathbf{O}_2, \mathbf{O}_3, \mathcal{Q}), \\
\mathbf{Y}_{\text{con}} &= \mathrm{Concat}(\mathbf{y}, \mathbf{y}, \mathbf{y}, \mathbf{y}_{\mathcal{Q}}),
\end{align}
where \(\mathbf{O}_i\) denote the expert outputs, and \(\mathbf{y}\) and \(\mathbf{y}_{\mathcal{Q}}\) are the labels for current batch and queue, respectively. Concatenating features along the batch dimension enlarges the pool of positive and negative pairs, improving contrastive learning.

We define two complementary losses: a supervised contrastive loss, which leverages label information to pull semantically similar samples closer together, and a self-supervised contrastive loss, which identifies positive pairs by mining class-consistent samples from both the current batch and the queue without relying on explicit labels:
\begin{equation}
\mathcal{L}_{\text{sup}} = \mathrm{SupCon}(\mathbf{O}_{\text{con}}, \mathbf{Y}_{\text{con}}),
\end{equation}
\begin{equation}
\mathcal{L}_{\text{self}} = \mathrm{SelfCon}(\mathbf{O}_{\text{con}}).
\end{equation}

The final objective is defined as a weighted sum of the two losses, controlled by a balancing factor \(\lambda\):
\begin{equation}
\mathcal{L}_{\text{mc}} = \mathcal{L}_{\text{sup}} + \lambda \mathcal{L}_{\text{self}},
\end{equation}
where \(\lambda\) controls the relative strength of the self-supervised signal. This unified contrastive framework jointly optimizes explicit semantic discrimination and implicit structural alignment, enhancing representation robustness and generalization in ultrasound standard plane recognition.

\noindent \textbf{MoE Recognition Branch.}
In standard plane recognition, expert subnetworks capture diverse feature patterns, but simple averaging may overlook sample-specific variations. To address this, we introduce a novel MoE classification branch equipped with a learnable sparse gating mechanism. Leveraging Gumbel-Softmax~\cite{lin2017feature}, the gate adaptively selects the most informative experts while remaining fully differentiable, improving robustness to anatomical ambiguity and imaging variability.

Let the semantic-structural feature be \(\mathbf{O} \in \mathbb{R}^{B\times C\times H\times W}\), which is fed into a lightweight gating network to generate expert logits \(\mathbf{l} \in \mathbb{R}^{B\times N}\). This gating network performs adaptive average pooling followed by a linear layer.  The Gating weights \(\mathbf{w}\) are then obtained using the Gumbel-Softmax function:
\begin{equation}
\mathbf{w} = \text{GumbelSoftmax}(\mathbf{l}, \tau),
\end{equation}
where \(\tau\) controls the sparsity of the distribution.. Given expert predictions \(\mathbf{z}_n \in \mathbb{R}^{B \times C}\), we stack them into \(\mathbf{Z} \in \mathbb{R}^{B \times N \times C}\) and compute the fused output as:
\begin{equation}
\mathbf{z}_{\text{fused}} = \sum_{n=1}^{N} w_n \cdot \mathbf{z}_n.
\end{equation}

The fused prediction \(\mathbf{z}_{\text{fused}}\) is supervised using a standard cross-entropy loss:
\begin{equation}
\mathcal{L}_{\text{moe}} = \text{CE}(\mathbf{z}_{\text{fused}}, \mathbf{y}).
\end{equation}

This design improves semantic classification flexibility and reduces redundant expert interaction, thus enhancing model generalization. To balance the main classification loss \(L_{\text{moe}}\) and the contrastive loss \(L_{\text{mc}}\), we employ a lightweight adaptive network. Given an input feature \(\mathbf{O} \in \mathbb{R}^{B\times C\times H\times W}\), it predicts a sample-specific weight \(\alpha = g(\mathbf{O}) \in (0,1)\). Overall, the total loss is computed as a weighted combination of the two losses:
\begin{equation}
L_{\text{total}} = \alpha \cdot L_{\text{moe}} + (1 - \alpha) \cdot L_{\text{mc}}.
\end{equation}

The balancing factor $\alpha$ is adaptively adjusted based on sample difficulty and the dynamics of the training process. This mechanism removes the need for manually tuned hyperparameters and enables end-to-end learning of optimal weights. Consequently, it enhances the stability and effectiveness of multi-task collaborative training.

\section{Experimental Results}

\subsection{Datasets and Evaluation Metrics}
\textbf{FPUS23}~\cite{prabakaran2023fpus23} is a public fetal ultrasound dataset for standard plane recognition, covering key anatomical views such as the head, abdomen, femur, and thorax, with expert annotations suitable for supervised learning. \textbf{CAMUS}~\cite{leclerc2019deep} is a cardiac ultrasound dataset originally designed for segmentation. we selected the apical two-chamber and four-chamber views and annotated them for classification. Its diversity across subjects makes it well suited for evaluating model generalization. Additionally, we use our in-house \textbf{LP2025} dataset, which contains clinically defined standard liver planes for abdominal ultrasound analysis. To fairly evaluate and compare our method with existing approaches, we adopt four commonly used metrics, including Accuracy, Precision, Recall, and F1-score.

\subsection{Implementation Details}
Our method is implemented using on the PyTorch framework and all experiments are conducted on a server equipped with an NVIDIA RTX 3090 GPU running a Python 3.8 environment. For data preprocessing, all images from the datasets are resized to $512 \times 512$, and several data augmentation techniques are applied, including random rotation, horizontal and vertical flipping, and brightness adjustment. The model is optimized using stochastic gradient descent (SGD) with a momentum of 0.9 and a weight decay of $1 \times 10^{-4}$. The initial learning rate is set to $1 \times 10^{-3}$ and decayed using a cosine annealing schedule. The batch size is 16, and training is conducted for up to 200 epochs.

\subsection{Comparison with State-of-the-Art Methods}
To evaluate the classification performance of our model, we conducted a comparative analysis with recent state-of-the-art methods. These include Diffmic~\cite{yang2023diffmicdualguidancediffusionnetwork}, which has demonstrated strong performance in medical image analysis; Area~\cite{chen2023area} and Shike~\cite{jin2023long}, which are CNN-based and show notable improvements; MetaFormer~\cite{yu2023metaformer} and Cast~\cite{ke2024learninghierarchicalimagesegmentation}, which are Transformer-based and achieve promising results; and SupMin~\cite{mildenberger2025taleclassesadaptingsupervised}, which incorporates improved supervised contrastive learning. For a fair comparison, all models are trained with consistent settings and evaluated under identical experimental conditions.

\begin{table}[ht]
\footnotesize
\centering
\setlength{\belowcaptionskip}{0cm}
\renewcommand{\arraystretch}{1.0}
\begin{centering}
	\setlength{\tabcolsep}{1.8mm}{
\begin{tabular}{lcccc}
\hline\specialrule{0.06em}{0pt}{1.9pt}
\multirow{2}{*}{Method}& \multicolumn{4}{c}{LP2025} \\ \cmidrule(lr){2-5}
& Accuracy$\uparrow$ & Precision$\uparrow$ & Recall$\uparrow$ & F1-score$\uparrow$ \\
\specialrule{0.02em}{0pt}{1.9pt}
Diffmic        & 80.07 & 75.35 & 77.81 & 76.27 \\
Area           & 80.39 & 75.43 & 76.21 & 75.04 \\
Shike          & 80.26 & 75.76 & 77.63 & 76.19 \\
Metaformer     & 80.13 & 77.10 & 77.00 & 76.44 \\
Cast           & 80.86 & 74.87 & 79.59 & 77.00 \\
Supmin         & 80.92 & 75.95 & 78.12 & 76.77 \\
Ours           & \textbf{82.30} & \textbf{78.11} & \textbf{80.92} & \textbf{79.32} \\
\hline\specialrule{0.06em}{0pt}{1.9pt}
\end{tabular}}

\caption{Quantitative comparisons of different models on the in-house LP2025 dataset.}
\label{tab:LP2025}
\end{centering}
\end{table}

\noindent \textbf{Results on FPUS23 Dataset.}
Table~\ref{tab:FPUS2023andCamus} shows that on the FPUS23 fetal standard plane dataset, our method achieves the highest performance across all metrics. Specifically, we obtain an Accuracy of 95.78\%, surpassing the second-best MetaFormer (95.52\%) by 0.26\%. Our method also achieves the highest F1-score, reaching 95.06\% and outperforming Area (94.40\%) and SupMin (94.34\%). These improvements stem from the proposed semantic-structure fusion module, which effectively captures both shallow structural cues and deep semantic features. Moreover, the cross-expert collaborative classification branch adaptively fuses predictions from multiple experts, thereby enhancing robustness and overall classification accuracy.

\subsubsection{Results on CAMUS Dataset.}
Table~\ref{tab:FPUS2023andCamus} shows that our SEMC framework consistently outperforms existing methods on the CAMUS cardiac standard plane dataset. SEMC achieves the highest Accuracy of 82.13\%, surpassing MetaFormer (81.52\%) and Area (81.59\%). It also obtains the best F1-score of 80.93\%, outperforming CAST (80.31\%) and SupMin (79.71\%). It demonstrate that SEMC addresses the large intra-class variability and high inter-class similarity characteristic of ultrasound imaging.
\begin{table}[htbp]
\footnotesize
\centering
\setlength{\belowcaptionskip}{0cm}
\renewcommand{\arraystretch}{1.0}
\begin{centering}
	\setlength{\tabcolsep}{2.8mm}{
\begin{tabular}{c c c c c}
\hline\specialrule{0.06em}{0pt}{1.9pt}
ACE & SAMC & $L_{\text{mc}}$ & Accuracy $\uparrow$ & F1-score $\uparrow$\\
\specialrule{0.02em}{0pt}{1.9pt}
\ding{55} & \ding{55} & \ding{55} & 80.26 & 76.98 \\
\ding{51} & \ding{55} & \ding{55} & 81.38 & 77.82 \\
\ding{51} & \ding{51} & \ding{55} & 81.51 & 77.91 \\
\ding{51} & \ding{55} & \ding{51} & 81.78 & 78.65 \\
\ding{51} & \ding{51} & \ding{51} & \textbf{82.30} & \textbf{79.32}\\
\hline\specialrule{0.06em}{0pt}{1.9pt}
\end{tabular}}

\caption{Ablation study of the proposed ACE, SAMC, and $L_{\text{mc}}$ on the LP2025 dataset.}
\label{tab:ablation_lp2025}
\end{centering}
\end{table}

\noindent \textbf{Results on LP2025 Dataset.}
Table~\ref{tab:LP2025} reports the results on the LP2025 dataset, where our method achieves the state-of-the-art performance. Specifically, it reaches an Accuracy of 82.30\%, outperforming Diffmic (80.07\%) by 2.23\%, and obtains an F1-score of 79.32\%, exceeding SupMin (76.77\%) by 2.55\%. These improvements highlight our model’s strong generalization in capturing discriminative structural differences and key semantic regions across diverse liver views, leading to superior classification performance.

\subsection{Ablation Study}
\noindent \textbf{Ablation Study of Each Component.}  
We conducted ablation studies on the LP2025 dataset to assess the contribution of each module in our framework. As shown in Table~\ref{tab:ablation_lp2025}, we progressively incorporated the SSFM and MCRM into the baseline. The SSFM includes the ACE and SAMC submodules, while MCRM introduces a contrastive loss, \textit{i.e.}, $L_{\text{mc}}$, to encourage expert branches to learn complementary representations. The baseline achieves 80.26\% accuracy and 76.98\% F1-score. Adding ACE increases accuracy to 81.38\% and F1-score to 77.82\%, demonstrating its effectiveness in capturing shallow structural cues. Besides, adding SAMC on top of ACE further enhances attention to spatial regions. Alternatively, ACE combined with $L_{\text{mc}}$ boosts accuracy to 81.78\% and F1 to 78.65\%, demonstrating the benefit of expert-guided contrastive learning. With all modules enabled, the model achieves the best performance, confirming that the three components are complementary and jointly enhance both feature representation and generalization.
\begin{figure}[t]
    \centering
    \includegraphics[width=\columnwidth, height=0.7\columnwidth]{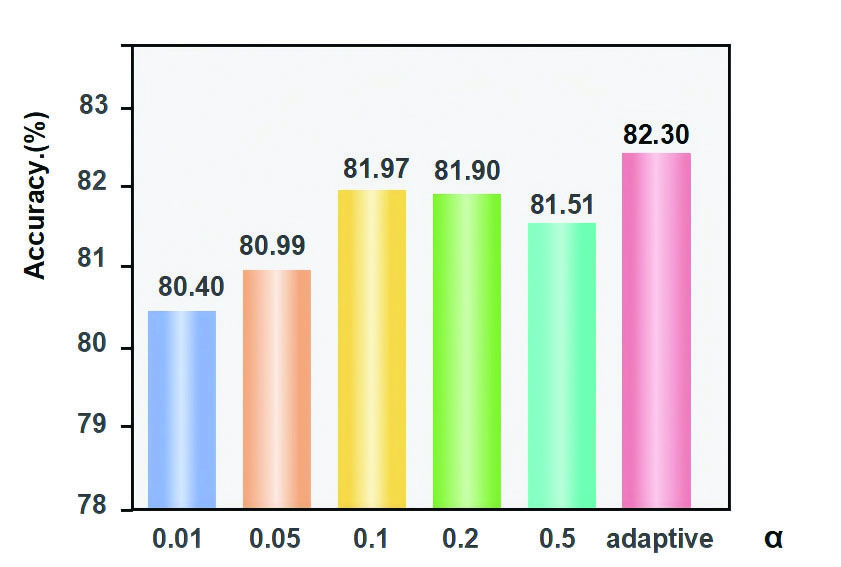} 
   
    \caption{Performance comparison for different values of the hyperparameter $\alpha$ on the LP2025 dataset.}

    \label{fig:ablation_alpha}
\end{figure}

\noindent \textbf{Ablation Study of Adaptive Parameter.}
Figure~\ref{fig:ablation_alpha} presents a sensitivity analysis of the hyperparameter \(\alpha\) defined in Equation~(19). We conduct systematic ablation experiments on the LP2025 dataset, testing fixed values \(\alpha\) of 0.01, 0.05, 0.1, 0.2, and 0.5, alongside our proposed adaptive coefficient. The results demonstrate that the adaptive strategy outperforms all fixed settings, significantly improving classification accuracy while greatly improving training stability and overall model robustness.

\section{Conclusion}
This paper presents a novel structure-enhanced mixture-of-experts contrastive learning framework, dubbed SEMC, for ultrasound standard plane recognition. Specifically, the proposed semantic-structure fusion module aligns and integrates shallow structural cues with deep semantic representations. Additionally, the mixture-of-experts contrastive recognition module is designed to specialize in different aspects of the feature space and collaboratively performs hierarchical contrastive learning, enabling the capture of fine-grained discriminative representations. More importantly, we establish a high-quality ultrasound dataset comprising six standard planes, addressing the scarcity of publicly available benchmarks. Extensive experiments on this in-house dataset and two public benchmarks demonstrate that SEMC consistently outperforms recent state-of-the-art methods.

\section{Acknowledgments}
This work was supported in part by the National Science Foundation of China under Grant62471448; in part by  Shandong Provincial Natural Science Foundation under Grant ZR2024YQ004; in part by TaiShan Scholars Youth Expert Program of Shandong Province under Grant No.tsqn202312109.

\bibliography{aaai2026}
\end{document}